\documentclass{Interspeech}

\interspeechcameraready

\title{Scaling and Prompting for Improved End-to-End Spoken Grammatical Error Correction}

\author[affiliation={}]{Mengjie}{Qian}
\author[affiliation={}]{Rao}{Ma}
\author[affiliation={}]{Stefano}{Bannò}
\author[affiliation={}]{Kate M.}{Knill}
\author[affiliation={}]{Mark J.F.}{Gales}

\affiliation{ALTA Institute, Department of Engineering}{University of Cambridge}{UK}

\email{\{mq227,rm2114,sb2549,kmk1001,mjfg100\}@cam.ac.uk}


\usepackage[table]{xcolor}
\usepackage{amsmath}
\usepackage{graphicx}
\usepackage[colorinlistoftodos]{todonotes}
\usepackage{float}
\usepackage{subcaption}
\usepackage{amsfonts}
\usepackage{multicol,multirow}
\usepackage[bottom]{footmisc}
\usepackage{booktabs, pifont}
\usepackage{enumitem}
\usepackage{soul}
\usepackage{tabularx, url}
\usepackage{comment,makecell}
\definecolor{gray}{rgb}{0.5,0.5,0.5}
\definecolor{Gray}{gray}{0.9}

\begin{document}

\maketitle

\footnotetext[1]{This paper reports on research supported by Cambridge University Press \& Assessment, a department of The Chancellor, Masters, and Scholars of the University of Cambridge.}

\begin{abstract}
Spoken Grammatical Error Correction (SGEC) and Feedback (SGECF) are crucial for second language learners, teachers and test takers. Traditional SGEC systems rely on a cascaded pipeline consisting of an ASR, a module for disfluency detection (DD) and removal and one for GEC. With the rise of end-to-end (E2E) speech foundation models, we investigate their effectiveness in SGEC and feedback generation. This work introduces a pseudo-labelling process to address the challenge of limited labelled data, expanding the training data size from 77 hours to approximately 2500 hours, leading to improved performance. Additionally, we prompt an E2E Whisper-based SGEC model with fluent transcriptions, showing a slight improvement in SGEC performance, with more significant gains in feedback generation. Finally, we assess the impact of increasing model size, revealing that while pseudo-labelled data does not yield performance gain for a larger Whisper model, training with prompts proves beneficial.

\end{abstract}


\section{Introduction}
Spoken Grammatical Error Correction (SGEC) has emerged as a critical task in the field of computer-assisted language learning (CALL), providing learners with essential feedback on their spoken language use. Unlike traditional text-based Grammatical Error Correction (GEC), which focuses on written content, SGEC must handle the complexities of spontaneous speech, including disfluencies (i.e., hesitations, repetitions, and false starts), accented speech, varied sentence structures and incomplete sentences commonly found in spoken language). These challenges make SGEC particularly demanding and call for innovative solutions in both model design and data handling.

Written GEC has a well-established research history~\cite{bryant2023}, with several shared tasks released in the past years~\cite{ng2014conll,bryant2019bea,multigec2024}. 
In contrast, SGEC remains relatively under-explored, with fewer datasets and methods dedicated to its unique challenges. 
Apart from the early pioneering work on manual transcriptions of second language (L2) Japanese learners of English by \cite{izumi2003automatic-full}, most research involving fully automated approaches in this area has emerged more recently, with progress driven by cascaded systems that combine automatic speech recognition (ASR), disfluency detection (DD), and grammatical error correction (GEC) modules~\cite{lu2020spoken,lu2022assessing, banno2023b_slate_grammatical, banno2024back}. These systems offer strong baseline performance but face challenges with error propagation across modules, limiting their overall effectiveness and robustness. The work of \cite{banno2024towards} presents a significant step forward by using Whisper~\cite{radford2023robust}, a speech foundation model, for end-to-end (E2E) spoken GEC and DD. Their approach reduces modular dependencies but highlights the need for more annotated training data for the E2E system to match cascaded performance in GEC tasks. 

In addition to providing learners with a grammatically corrected transcription, it is essential to offer meaningful feedback that helps them understand their mistakes rather than simply presenting a `ready-made’ correction. Feedback is a crucial component in CALL applications, offering learners actionable insights into where and how they have made errors. Effective feedback must be easy to understand, informative, and supportive of language learning. Therefore, in contrast to SGEC, which aims to correct grammatical errors, SGEC feedback (SGECF) aims to deliver more detailed guidance by not only highlighting errors but also explaining why they occurred and how learners can improve.  An interesting early approach \cite{lee2014grammatical} proposed a feedback system using a statistical model for grammatical error detection and feedback in spoken language. The authors of \cite{banno2024towards} also addressed the challenge of providing accurate grammatical feedback through an E2E model, although their approach did not yield significant performance improvements. While other works have focused on grammatical feedback comment generation for writing, particularly with the advent of LLMs~\cite{fei-etal-2023-enhancing, kaneko-okazaki-2024-controlled, song-etal-2024-gee}, grammatical feedback for speaking remains largely unexplored. 

This work builds upon previous research in SGEC, exploring novel methods to enhance both SGEC and SGEC feedback performance.
A major challenge in SGEC advancement is the limited availability of high-quality annotated spoken datasets, though initiatives like the Speak \& Improve Corpus~\cite{qian2024speak, knill2024speak} are beginning to address this gap. Meanwhile, we propose a pseudo-labelling process to leverage abundant audio data for SGEC training. 
To generate feedback on edits, GEC transcriptions are compared with fluent transcriptions. Therefore, we propose to prompt the model with fluent transcriptions to provide additional information, enhancing SGEC performance. Both approaches, pseudo-labelling and prompting with fluent transcriptions, lead to improvements in SGEC and feedback performance, surpassing a cascaded system.

\section{Method}
\label{sec:method}
\subsection{Cascaded System}
\label{sec:cascaded}
\begin{figure}
    \centering
    \includegraphics[width=0.9\linewidth]{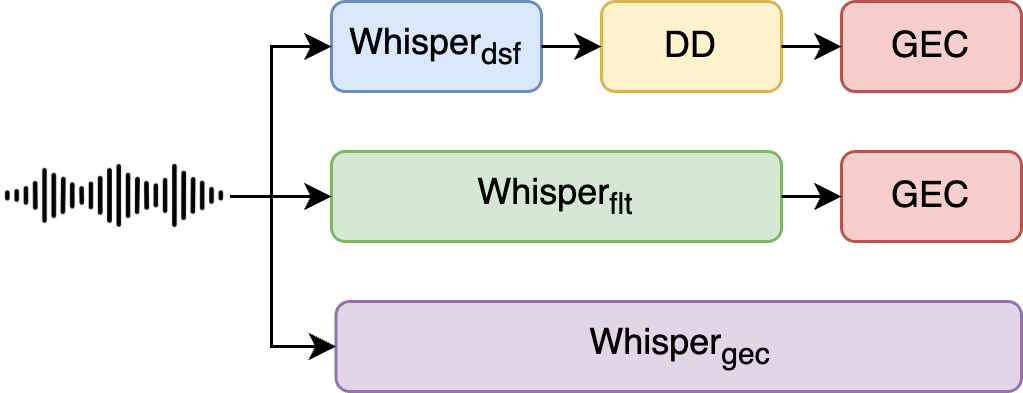}
    \caption{Illustration of the E2E SGEC and cascaded systems.}
    \label{fig:sgec}
\end{figure}

A traditional cascaded spoken GEC system consists of three distinct modules: ASR, DD and GEC (Figure~\ref{fig:sgec}). First, the ASR module transcribes the speech into text. Then, the DD module identifies and removes disfluencies, such as interruptions, repetitions and hesitations, from the text transcription. Finally, the GEC module corrects grammatical errors in the transcribed speech, producing grammatically correct transcriptions. This modular approach combines speech recognition data, disfluency detection data, and text-based GEC data, all of which are more readily available than annotated spoken GEC data, helping to address the challenge of limited annotated spoken GEC data. 

Previous work~\cite{banno2024towards} introduced an end-to-end DD model using Whisper, referred to as Whisper$_\text{flt}$. This architecture integrates the traditional ASR and DD, generating fluent transcriptions from spoken audio that may contain disfluencies. When combined with GEC, the Whisper$_\text{flt}$ + GEC cascaded system outperforms the traditional modular spoken GEC system, which relies on three separate modules. This architecture serves as the baseline cascaded system in this work.

\subsection{End-to-end System}
\label{sec:e2e}
Recent advancements in foundation speech models, such as Whisper~\cite{radford2023robust}, trained on over 680 thousand hours of labelled data across 100 languages using a multi-task learning approach, have gained popularity.
This training setup enables Whisper to be adapted for tasks beyond its initial capabilities, including speech recognition for unseen languages~\cite{qian2024learn,timmel2024fine}, speech translation across various language pairs~\cite{ma2025cross,cheng2024task}, and other spoken language understanding tasks beyond ASR~\cite{ma2024investigating,goron2024improving}.

In this work, we extend Whisper for E2E spoken grammatical error correction by fine-tuning it on grammatically corrected transcriptions (Whisper$_\text{gec}$). The model directly generates grammatically corrected transcriptions from spoken input, eliminating the need for separate modules. While prior work has explored leveraging Whisper for spoken GEC~\cite{banno2024towards}, our approach introduces several novel methods, including pseudo-labelling with unlabelled data and model prompting. These methods lead to improvements in both SGEC and feedback performance, surpassing cascaded systems.

\subsection{Pseudo-labelling Process}
\label{sec:pseudo-label}
While annotated spoken GEC data is limited, audio recordings are widely available. A common challenge for E2E models is their need for large amounts of training data to be effective. To address this and increase the training data size for SGEC, we propose a fully automated labelling process to generate pseudo-GEC transcriptions for audio data. This approach leverages the vast amount of readily available audio data, significantly expanding the training data for SGEC model development. 

Specifically, we utilise a cascaded GEC system for the labelling process. Below are the detailed steps: 
\begin{itemize}
    \item Step1: Generate automatic disfluent transcriptions for the audios using a Whisper model. The model employed here is Whisper small.en, fine-tuned on 20 hours of Linguaskill~\cite{ludlow2020official} data with segment-level timestamp information and truecasing. Timestamp information is generated from forced alignment using HTK Hidden Markov Model (HMM)-Gaussian Mixture Model (GMM) MPE L2 English models. Truecasing is applied by capitalising the first character of each sentence, based on the manual transcriptions.
    \item Step2: Segment the unlabelled audio data into short  segments based on the automatic disfluent transcriptions from Step1, using punctuation marks (full stops, question marks, and exclamation marks) to identify phrase boundaries.
    \item Step3: Decode the segmented audio to obtain automatic fluent transcriptions using the Whisper$_\text{flt}$ model from~\cite{banno2024towards}.
    \item Step4: Apply a text-based GEC system to the fluent transcriptions from Step3 to generate phrase-level GEC transcriptions. The GEC system uses the same setup from~\cite{banno2024towards}.
\end{itemize}

With this process, we annotated around 2500 hours of audio data, collected from the Speaking section of Linguaskill.

\subsection{Prompting Whisper}
\label{sec:prompt}
\begin{figure}[!htbp]
    \centering
    \includegraphics[trim=0 4mm 0 6mm, width=1\linewidth]{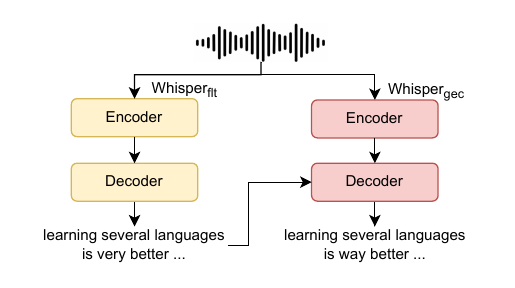}
    \caption{The E2E SGEC system prompting with additional ASR transcriptions.}
    \label{fig:whisper-prompt}
\end{figure}
\noindent
To generate grammatically correct transcriptions, the model can benefit from additional contextual information. Our approach builds on this idea by prompting the model with fluent transcriptions, which have disfluencies removed. This extra guidance helps the model better understand the structure of the spoken language. 
Specifically, we fine-tune a Whisper model on a dataset that includes both fluent transcriptions and their corresponding speech input (as illustrated in Figure~\ref{fig:whisper-prompt}). The fluent transcriptions, generated by a Whisper model fine-tuned on fluent transcription (Whisper$_\text{flt}$), provide clearer context for learning grammatical corrections. This method enables the model to leverage fluent transcriptions without relying solely on the GEC transcriptions, helping it to focus on language structure and improving its ability to generate accurate GEC transcriptions and provide useful feedback.

\section{Experimental Setup}
\subsection{Datasets}
\label{sec:dataset}
This paper uses Linguaskill~\cite{ludlow2020official} labelled and unlabelled training sets to build systems, Linguaskill dev set to select hyperparameters, and Linguaskill test set and Speak \& Improve Corpus~\cite{knill2024speakimprovecorpus} dev set for system evaluation.

\noindent \textbf{Linguaskill:} The data used in our study are obtained from candidate responses to the Speaking module of the Linguaskill tests for L2 learners of English, provided by Cambridge University Press \& Assessment~\cite{ludlow2020official}.
The dataset is gender-balanced and includes approximately 30 different L1s, with proficiency levels spanning A2 to C according to the Common European Framework of Reference (CEFR)~\cite{cefr2001}. A subset of the dataset is manually labelled (LNG$_\text{lbl}$), while the majority remains unlabelled (LNG$_\text{unl}$).
LNG$_\text{lbl}$ has been annotated with information on disfluencies and grammatical error corrections~\cite{knill23_slate-full}. Since responses can last up to 60 seconds, they were segmented into `sentences' through automatic time alignment based on manually marked boundaries between speech phrases.

\noindent \textbf{S\&I:} The Speak \& Improve (S\&I) Corpus 2025~\cite{knill2024speakimprovecorpus} is a dataset of L2 learner speech created to support research in spoken language assessment and feedback. Drawn from recordings on the S\&I version 1 platform spanning 2019 to 2024~\cite{nicholls23_interspeech}, the corpus offers diverse learner audio recordings, manual transcriptions, disfluency annotations, grammatically corrected transcripts, and associated CEFR proficiency scores from A2 to C. 

Further details about the data can be found in Table \ref{T:data_stats}.

\begin{table}[!htbp]
\footnotesize
    \centering
    \caption{Statistics of datasets.}
    \label{T:data_stats}
    \begin{tabular}{l|c|c|c|c|c}
    \toprule
    Corpus &  Split & Hours & Speakers & Utts/Sents & Words \\
    \midrule
    \multirow{3}*{LNG$_\text{lbl}$} & train & 77.6 & 1,908 & 34,790 & 502K \\
    & dev & 7.8 & 176 & 3,347 & 49K \\
    & test & 11.0 & 271 & 4,565 & 69K \\
    \cmidrule{1-6}
    LNG$_\text{unl}$ & train & 2521.6 & - & 708,613 & 15M \\
    \cmidrule{1-6}
    \multirow{1}*{S\&I} & dev & 20.8  & - & 2,866  & 105k  \\
    \bottomrule
    \end{tabular}
\end{table}

\subsection{Model Setup}
\label{sec:model_setup}
Whisper is used to train E2E models in this work, specifically Whisper$_\text{flt}$ with fluent references and Whisper$_\text{gec}$ with grammatically correct references. The small.en and large-v2 versions serve as the foundation models in this paper. 
The pre-trained models are fine-tuned on the Linguaskill training set using different manual references (fluent or GEC). The small.en model is trained for 30,000 steps with a batch size of 5. The large-v2 model is trained for 2 epochs with a batch size of 1 and a gradient accumulation step of 8. The learning rate is initialised to 1e-6, with linear decay applied during training. Beam search with a width of 5 is used during decoding.

For the baseline cascaded SGEC system, a text-based GEC is used. The system is initialised from the BART model~\cite{bart2020} provided by the HuggingFace Transformer Library \cite{huggingface} (\emph{facebook/bart-base}). The model is trained on the EFCAMDAT and BEA-2019 data for 19 epochs with a maximum sequence length of 256, a batch size of 16, a gradient accumulation step of 4, and a learning rate of 2e-6. It is then further fine-tuned on the Linguaskill data for 5 epochs with the encoder frozen, and the learning rate is reduced to 1e-5.

\subsection{Evaluation Metrics}
\label{sec:evaluation_metric}
Evaluating spoken GEC is challenging. Previous studies~\cite{banno2024towards,lu2022assessing} have demonstrated that both Translation Edit Rate (TER) and Word Error Rate (WER) are relevant metrics for spoken GEC. Both metrics report similar trends in evaluating spoken GEC, making it unnecessary to use both. In this work, we adopt WER as the primary metric for its simplicity and clarity.

To assess SGEC feedback performance, we use MaxMatch ($M^2$)~\cite{dahlmeier2012better} to capture phrase-level edits, using $M^2$ from fluent and GEC manual transcriptions as references, and $M^2$ from machine-generated fluent and GEC transcriptions as predictions. ERRANT~\cite{bryant2017automatic} is then used to compute Precision, Recall, and F$_{0.5}$ scores. We opt for F$_{0.5}$ to emphasise precision, which is critical for feedback generation and essential for maintaining user trust, as highlighted in the CoNLL-2014 Shared Task~\cite{ng2014conll}.


\section{Experiments}
\label{sec:exp}
\subsection{Scaling Training Data Using Pseudo-labelling}
\label{sec:exp_pseudo}
Previous work~\cite{banno2024towards} demonstrated promising results with the E2E Whisper$_{\text{gec}}$, achieving a WER of 13.49\% on the LNG$_{\text{lbl}}$ test set. However, a performance gap remains compared to the cascaded system, which achieves a lower WER of 12.96\%~\cite{banno2024towards}. In this experiment, we replicate their models and investigate whether incorporating pseudo-labelled data into the training process can bridge the gap between the E2E and cascaded systems. The models are evaluated on both the LNG$_{\text{lbl}}$ test set and open-source S\&I dev set.
The replicated models show results consistent with~\cite{banno2024towards}. As shown in Table~\ref{tab:results_pseudo}, the cascaded system with  Whisper$_{\text{flt}}$ (small.en) and a text-based GEC achieves a WER of 13.24\% on LNG$_{\text{lbl}}$ and 16.91\% on S\&I. In comparison, the E2E Whisper$_{\text{gec}}$ model achieves a WER of 13.48\% and 17.76\% on LNG$_{\text{lbl}}$ and S\&I, respectively. As in \cite{banno2024towards}, the cascaded system outperforms the E2E model.
When fine-tuning Whisper$_{\text{gec}}$ with only pseudo-labelled data (LNG$_{\text{unl}}$), the model achieves a WER of 14.16\%, just 5\% worse than the model trained on labelled data. This shows the potential of pseudo-labelled data in improving E2E SGEC. Further fine-tuning with LNG$_\text{lbl}$ after training with LNG$_\text{unl}$ significantly boosts performance, reducing the WER to 12.72\% and outperforming the cascaded system by 4.0\% relatively. However, this model only outperforms the cascaded system by 0.07\% on the S\&I dev set.

Increasing the model size to large-v2 improves performance on both cascaded and E2E models. The cascaded system reduces WER by 10.0\% on LNG$_\text{lbl}$ and 17.3\% on S\&I, while Whisper$_{\text{gec}}$ (large-v2) shows even greater improvements, with a 17.7\% reduction on LNG$_\text{lbl}$ and 25.5\% on S\&I, outperforming the cascaded system. These results suggest that larger models are better at leveraging labelled data compared to smaller models.
However, pseudo-labeled data does not show the same effectiveness with the large-v2 model. 
One possible reason is that the pseudo-GEC transcriptions are generated using the small.en model, which is smaller than Whisper$_{\text{gec}}$ large-v2 (see details in Section~\ref{sec:pseudo-label}). The size mismatch and potentially lower transcription quality from the small.en model likely reduces the effectiveness of pseudo-labeled data for the large-v2 model.


\begin{table}[!htbp]
    \centering
    \caption{Evaluation (WER) of Whisper$_{\text{gec}}$ performance with pseudo-labelled data on LNG$_{\text{lbl}}$ test and S\&I dev sets. Models are fine-tuned from the Whisper small.en and large-v2 models.}
    \label{tab:results_pseudo}
    \begin{tabular}{@{}c|l@{ }|c|c|c|c@{ }}
    \toprule
    \multicolumn{1}{@{ }c|}{\multirow{2}*{Model}} & \multirow{2}*{FT (cont.)} & \multicolumn{2}{c|}{small.en} & \multicolumn{2}{c}{large-v2}\\
    & & LNG$_{\text{lbl}}$ & S\&I & LNG$_{\text{lbl}}$ & S\&I \\
    \midrule
    \multicolumn{2}{c|}{Whisper$_{\text{flt}}$ + GEC} & 13.24 & 16.91 & 11.81 & 13.99 \\
    \cmidrule{1-6}
    \multirow{3}*{Whisper$_{\text{gec}}$}  & LNG$_{\text{lbl}}$ & 13.48 & 17.76 & \textbf{11.10} &  \textbf{13.21} \\
        & LNG$_{\text{unl}}$ & 14.16 & 18.11 & 12.93 & 15.92\\
        & \hspace{1mm} + LNG$_{\text{lbl}}$ & \textbf{12.72} & 16.84 & \textbf{11.10} & 13.93\\
    \bottomrule
    \end{tabular}
\end{table}

\subsection{Prompting with Additional Information}
\label{sec:exp_prompt}

\begin{table}[!t]
    \centering
    \caption{Evaluation (WER) of Whisper$_{\text{gec}}$ performance with different text prompts on LNG$_{\text{lbl}}$ test and S\&I dev sets. $\dagger$ indicates the improvement over the cascaded system is statistically significant with $p < 0.001$.}
    \label{tab:whisper_prompt}
    \begin{tabular}{@{ }l@{ }|@{ }l@{ }|p{6mm}p{6mm}|p{6mm}p{6mm}}
    \toprule
    \multirow{2}{*}{Model Name} & \multirow{2}{*}{Prompt} & \multicolumn{2}{c|}{small.en} & \multicolumn{2}{c}{large-v2} \\
    & & LNG$_{\text{lbl}}$  & S\&I & LNG$_{\text{lbl}}$  & S\&I \\
    \midrule
     \multicolumn{2}{c|}{Whs$_\text{flt}$ + GEC} & \multicolumn{1}{c}{13.24} & 16.91 & 11.81 & 13.99\\
    \cmidrule{1-6}
    Whs$_\text{gec}$ & -  &  13.48 & 17.76 & 11.10$^\dagger$ & 13.21$^\dagger$\\
    Whs$_\text{gec+text-flt}$       &  Whs$_\text{flt}$  &  13.32 & 17.28 &  11.08$^\dagger$ & 13.09$^\dagger$ \\
    Whs$_\text{gec+text-flt-SA}$    & Whs$_\text{flt-SA}$&  13.21 &  17.17 &  11.04$^\dagger$ & \textbf{13.08}$^\dagger$ \\
    \cmidrule{1-6}
    Whs$_\text{gec+text-flt}$ (init)   &  Whs$_\text{flt}$  & \textbf{12.80} & \textbf{16.78}  & \textbf{10.93}$^\dagger$ & 13.38$^\dagger$\\
    \bottomrule
    \end{tabular}
\end{table}

\noindent
In this experiment, we investigate whether prompting Whisper with additional information can enhance SGEC performance. 
First, fluent transcriptions for the LNG$_\text{lbl}$ training set are generated by removing disfluencies using an E2E model (Whisper$_\text{flt}$). Whisper$_\text{gec}$ is then trained with audio as input, GEC transcriptions as reference, and Whisper$_\text{flt}$ transcriptions as the prompt. This model (Whisper$_\text{gec+text-flt}$) achieves 13.32\% WER on the LNG$_{\text{lbl}}$ test set and 17.28\% on the S\&I dev set, slightly outperforming the non-prompted Whisper$_\text{gec}$ model (Table~\ref{tab:whisper_prompt}).
Since Whisper$_\text{flt}$ performs better on the LNG$_\text{lbl}$ training set (as it's trained on this dataset), there is a slight mismatch in fluent transcriptions during training and inference for Whisper$_\text{gec+text-flt}$. 
To address this, SpecAugment~\cite{park19specaug} is applied during Whisper$_\text{flt}$ decoding on the training set to align the WER on the training set with that of the dev set. Specifically, two frequency masks ($F=22$), two time masks ($T=50$), and time warping ($W=5$) are used on the training speech. Fluent transcriptions are generated from this perturbed dataset. 
Prompting Whisper$_\text{gec}$ with these transcriptions (Whisper$_\text{flt-SA}$) yields 13.21\% WER on LNG$_{\text{lbl}}$ and 17.17\% on S\&I, showing further improvement. 
However, gains on LNG$_\text{lbl}$ remain marginal, and S\&I performance still lags behind the Whisper$_\text{flt}$ + GEC system.

Building on the potential of pseudo-labels, as shown in Section~\ref{sec:exp_pseudo}, we assess its effectiveness when combined with model prompting. Initialising Whisper$_\text{gec}$ with pseudo-labelled data, followed by fine-tuning with labelled data and fluent transcriptions as prompts (Whisper$_\text{gec+text-flt}$ (init)), improves performance on both LNG$_{\text{lbl}}$ and S\&I. This model outperforms the cascaded system by 3.3\% on LNG$_{\text{lbl}}$ and 0.7\% on S\&I, making it the best-performing model based on the small.en version.

Increasing model size does not reduce the benefits of prompting.
While the Whisper$_\text{gec}$ large-v2 model outperforms the cascaded Whisper$_\text{flt}$ (large-v2) + GEC system, training with prompts further improves results. Whisper$_\text{gec+text-flt-SA}$ reduces the WER to 11.04\% on LNG$_{\text{lbl}}$ and 13.08\% on S\&I.
However, initialising the large-v2 model with pseudo-labelled data yields inconsistent results on LNG$_{\text{lbl}}$ and S\&I, unlike the small.en model. This discrepancy is likely due to the use of small.en model in the pseudo-labelling process, leading to compromised GEC transcription quality for the large-v2 model.

\subsection{Analysis on Feedback}
\label{sec:exp_feedback}

\noindent
Feedback is a critical aspect to evaluate. With improved SGEC performance using model prompting, larger model size and pseudo-labelled data, we assess their impact on feedback performance. Here, we focus on the large-v2 models as they consistently outperform the small.en model. Table~\ref{tab:results_feedback} presents the SGEC feedback performance for various GEC models based on large-v2, evaluated on the LNG$_{\text{lbl}}$ test and S\&I dev sets. 
Previous work highlighted a significant feedback gap between the E2E SGEC model and the cascaded system. The Whisper$_\text{gec}$ (small.en) model achieved an F\textsubscript{0.5} of 26.40, compared to 39.74 for the Whisper$_\text{flt}$ (small.en) + GEC system~\cite{banno2024towards}.
With the large-v2 model, this gap narrows, reducing the F\textsubscript{0.5} difference between Whisper$_\text{gec}$ and the cascaded system from 13.34 to 2.17. Model prompting, proven effective in SGEC (Section~\ref{sec:exp_prompt}), also improves feedback performance. 
Although the improvement in WER from prompting is modest, the Whisper$_\text{gec}$ model with prompting (Whisper$_\text{gec+text-flt}$) closely matches the cascaded system in feedback, reducing the F\textsubscript{0.5} gap from 2.78 to -0.57 on the LNG${_\text{lbl}}$ test set and from 3.35 to 0.48 on S\&I. 
Prompting with SpecAugment applied or training from a model initialised from pseudo-labelled data yields similar performance to Whisper$_\text{gec+text-flt}$. We also explored using GPT-4o to correct grammar errors in the fluent transcriptions from Whisper$_\text{flt}$, but this did not improve GEC or feedback performance.

\begin{figure}[!htbp]
    \centering
    \includegraphics[trim=0 8mm 0 2mm, width=1.0\linewidth]{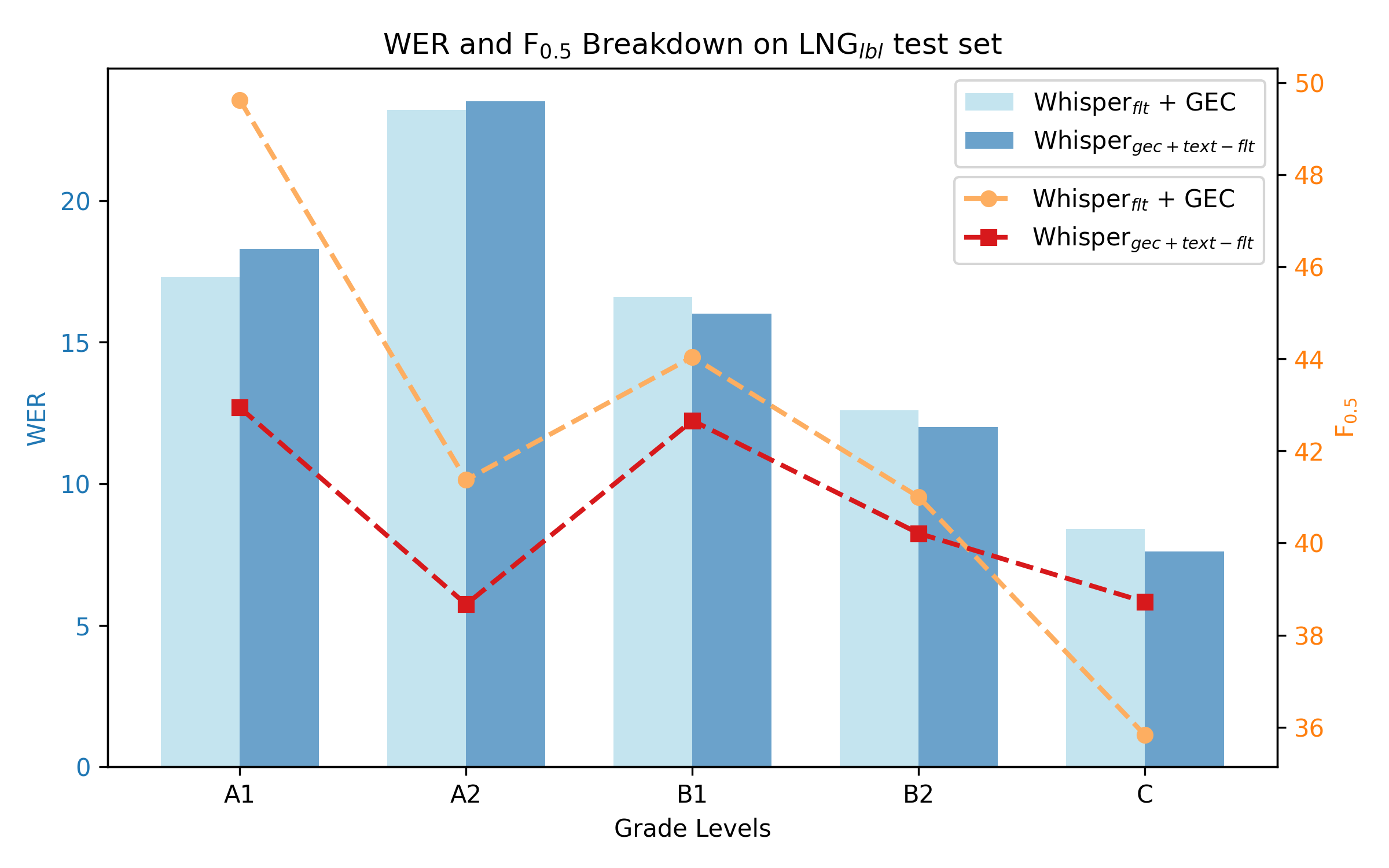}
    \caption{WER and F$_\text{0.5}$ breakdown on LNG$_\text{lbl}$ test set.}
    \label{fig:breakdown_lng}
\end{figure}

Figure~\ref{fig:breakdown_lng} shows performance breakdown by grade levels. Whisper$_\text{gec+text-flt}$ outperforms the cascaded system for the LNG$_\text{lbl}$ test set at levels B1, B2 and C. However, feedback remains more challenging for the E2E model, with Whisper$_\text{gec+text-flt}$ only outperforming the cascaded system at level C.


\begin{table}[!t]
    \centering
    \caption{Feedback evaluation of LNG$_{\text{lbl}}$ test and S\&I dev sets on various GEC models, with performance evaluated against fluent transcriptions generated from the Whisper$_\text{flt}$ model.}
    \label{tab:results_feedback}
    \begin{tabular}{@{ }l|p{6mm}p{6mm}p{6mm}|p{6mm}p{6mm}p{6mm}}
    \toprule
    \multirow{2}{*}{GEC Model} & \multicolumn{3}{c|}{LNG$_\text{lbl}$} & \multicolumn{3}{c}{S\&I} \\
     &  P & R & F\textsubscript{0.5} &  P & R & F\textsubscript{0.5}\\
    \midrule
    Whs$_\text{flt}$+GEC  & \textbf{46.60} & 26.61 & 40.51 & \textbf{49.43} & 28.51 & \textbf{43.10}\\
        \cmidrule{1-7}
        Whs$_\text{gec}$  & 40.38 & 31.91 & 38.34 & 41.87 & 33.06 & 39.75\\
        Whs$_\text{gec+text-flt}$ & 43.92 & \textbf{32.63} & \textbf{41.08} & 45.56 & \textbf{33.88} & 42.62   \\
    \bottomrule
    \end{tabular}
\end{table}

\section{Conclusions}
In this work, we explore the use of pseudo-labelled GEC data to scale up the training size of an E2E SGEC model, demonstrating its effectiveness in enhancing the model's performance. Increasing the model size also improves the capability of an E2E model for both SGEC and feedback tasks. Additionally, incorporating extra information through model prompting during training provides further improvements, even in larger models. Prompting the large-v2 Whisper$_\text{gec}$ model with fluent transcriptions achieves the best SGEC performance, with a WER of 11.08\%. This model also achieves a F\textsubscript{0.5} of 41.08 on LNG$_\text{lbl}$ test and 42.62 on S\&I dev for feedback performance, closely matching the best cascaded system. These results highlight the combined effectiveness of pseudo-labeling, model size scaling, and prompting in improving both SGEC and feedback tasks.


\clearpage
\bibliographystyle{IEEEtran}
\bibliography{mybib}

\end{document}